\documentclass[11pt]{article}
\usepackage{threeparttable}
\usepackage[
singlelinecheck=false % <-- important
]{caption}
\usepackage{coling2020}
\usepackage{times}
\usepackage{url}
\usepackage{latexsym}
\usepackage{multirow}
\usepackage{multicol}
\usepackage{booktabs}
\usepackage{graphicx}
\usepackage{caption}
\usepackage{xcolor}
    \usepackage{hyperref}
\newcommand{\header}[1]{\vspace{1mm}\noindent\textbf{#1}.}

\title{Optimizing Transformer for Low-Resource Neural Machine Translation}

\author{Ali Araabi \\
  Informatics Institute \\
  University of Amsterdam \\
  {\tt a.araabi@uva.nl} \\\And
  Christof Monz \\
  Informatics Institute \\
  University of Amsterdam \\
  {\tt c.monz@uva.nl} \\}

\date{}

\begin{document}

\maketitle
%\vspace{-2cm}
\begin{abstract}

Language pairs with limited amounts of parallel data, also known as low-resource languages, remain a challenge for neural machine translation. 
While the Transformer model has achieved significant improvements for many language pairs and has become the de facto mainstream architecture, its capability under low-resource conditions has not been fully investigated yet. 
Our experiments on different subsets of the IWSLT14 training data show that the effectiveness of Transformer under low-resource conditions is highly dependent on the hyper-parameter settings. % as well as the size of subword vocabularies. 
Our experiments show that using an optimized Transformer for low-resource conditions improves the translation quality up to 7.3 BLEU points compared to using the Transformer default settings.
  
\end{abstract}

\section{Introduction}
\label{intro}
Despite the success of Neural Machine Translation~(NMT)~\cite{sutskever2014sequence,cho2014learning,bahdanau2014neural}, for the vast majority of language pairs for which only limited amounts of training data exist (a.k.a. low-resource languages), the performance of NMT systems is relatively poor~\cite{koehn2017six,GuHDL18}.
Most approaches focus on exploiting additional data to address this problem~\cite{gulcehre2015using,sennrich2015improving,he2016dual,FadaeeBM17a}. 
However, \newcite{sennrich2019revisiting} show that a well-optimized NMT system can perform relatively well under low-resource data conditions.
%even without involving additional data.
Unfortunately, their results are confined to a recurrent NMT architecture \cite{sennrich-etal-2017-nematus}, and it is not clear to what extent these findings also hold for the nowadays much more commonly used Transformer architecture \cite{vaswani2017attention}.
%, which generally outperforms recurrent architectures and has become the mainstream model in NMT. 

Like all NMT models, Transformer requires setting various hyper-parameters but researchers often stick to the default parameters, even when their data conditions differ substantially from the original data conditions used to determine those default values \cite{GuWCLC18,AharoniJF19}. 

In this paper, we explore to what extent hyper-parameter optimization, which has been applied successfully to recurrent NMT models for low-resource translation, is also beneficial for the Transformer model. 
%We first explore the effect of different approaches to extract subword units on translation quality. %
% We show that with the appropriate settings, ranging from the number of BPE merge operations to the degree of label smoothing,
%
We show that with the appropriate settings, including the number of BPE merge operations, attention heads, and layers up to the degree of dropout and label smoothing,
%
% to the degree of label smoothing,
%
%, along with adopting a few regularization techniques, 
translation performance can be increased substantially, even for data sets with as little as 5k sentence pairs. 
% This even holds for very limited amounts of parallel data, with as little as 5k sentence pairs. 
Our experiments on different corpus sizes, ranging from 5k to 165k sentence pairs, show the importance of choosing the optimal settings with respect to data size.

%Finally, we observe that unlike Transformer base model that cannot outperform the optimized RNN system on small datasets, the optimized Transformer performs substantially better in terms of BLEU score.

\blfootnote{
    %
    % for review submission
    %
    %
    % % final paper: en-uk version 
    %
     \hspace{-0.65cm}  % space normally used by the marker
     This work is licensed under a Creative Commons  Attribution 4.0 International Licence. Licence details:
     \url{http://creativecommons.org/licenses/by/4.0/}.
    % 
    % % final paper: en-us version 
    %
    % \hspace{-0.65cm}  % space normally used by the marker
    % This work is licensed under a Creative Commons 
    % Attribution 4.0 International License.
    % License details:
    % \url{http://creativecommons.org/licenses/by/4.0/}.
}
\section{Hyper-Parameter Exploration}
In this section, we first discuss the importance of choosing an appropriate degree of subword segmentation before we describe the other optimal hyper-parameter settings.

\header{Vocabulary representation}
\label{low-resource-bpe}
In order to improve the translation of rare words, word segmentation approaches such as Byte-Pair-Encoding (BPE)~\cite{sennrich2015improving} have become standard practice in NMT. 
This is especially true for language pairs with small amounts of data where rare words are a common phenomenon. \newcite{sennrich2019revisiting} show that reducing the number of BPE merge operations can result in substantial improvements of up to 5 BLEU points for a recurrent NMT model. 
It is natural to assume that reducing the BPE vocabulary is similarly effective for Transfomer.

\header{Architecture tuning}
A current observation in neural networks, and in particular in Transformer architectures, is that increasing the number of model parameters improves performance~\cite{raffel2019exploring,wang2019learning}. 
However, those findings are mostly obtained for scenarios with ample training data and it is not clear if they are directly applicable to low-resource conditions. While \newcite{abs-2004-04418} show that using fewer Transformer layers improves the quality of low-resource NMT, we expand our exploration towards
the effects of using a narrow and shallow Transformer by reducing i) the number of layers in both the encoder and decoder, ii) the number of attention heads, iii)~feed-forward layer dimension~($d_{\mathit{ff}}$), and iv) embedding dimensions~($d_{\mathit{model}}$). 

\header{Regularization}
Following \newcite{sennrich2019revisiting}, we analyze the impact of regularization by applying  dropouts to various Transformer components \cite{bouthillier2015dropout}.
In addition to regular dropout which is applied to the output of each sub-layer (feed-forward and self-attention) and after adding the positional embedding in both encoder and decoder~\cite{vaswani2017attention}, we employ attention dropout after the softmax for self-attention and also activation dropout inside the feed-forward sub-layers. Moreover, we drop entire layers using layer dropout~\cite{fan2019reducing}. We further drop words in the embedding matrix using discrete word dropout~\cite{gal2016theoretically}. 
We also experiment with larger label-smoothing factors \cite{muller2019does}.

\section{Experiments}

\begin{table}[t]
	\centering
%	  \begin{threeparttable}

	\begin{tabular}{r l l }
	
		\toprule
		
		\multicolumn{1}{l}{Step} & \multicolumn{1}{l}{Hyper-parameter} & \multicolumn{1}{c}{Values}  \\ 
		\midrule
		1 &  feed-forward dimension  & $128$, $256$, $512$, $1024$, \underline{$2048$}, $4096$\\
		2 & embedding dimension & $256$, \underline{$512$}, $1024$\\
		3&  attention heads & $1$, $2$, $4$, \underline{$8$}, $16$\\
		4& dropout  & \underline{$0.1$}, $0.2$, $0.3$, $0.4$, $0.5$\\
		5& number of layers &$1$, $2$, $3$, $4$, $5$, \underline{$6$}, $7$	\\
		6 &  label smoothing   & \underline{$0.1$}, $0.2$, $0.3$, $0.4$, $0.5$, $0.6$, $0.7$, $0.8$\\
		7 & enc/dec layer dropout & \underline{$0$}, $0.1$, $0.2$, $0.3$, $0.4$\\
		8 &  src/tgt  word dropout & \underline{$0$}, $0.1$, $0.2$, $0.3$ \\
		9 & attention dropout & \underline{$0$}, $0.1$, $0.2$, $0.3$ \\
		10 &  activation dropout  & \underline{$0$}, $0.1$, $0.2$, $0.3$, $0.4$, $0.5$ \\
		11 & embedding layer normalization & yes, \underline{no}  \\
		12 & batch size & $512$, $1024$, $2048$, \underline{$4096$}, $8192$ , $12288$ \\
		13 & learning rate scheduler & 
		\underline{Transformer standard}, inverse square root\\
		14&warm-up steps &$2000$, \underline{$4000$}, $5000$, $6000$, $8000$, $10000$ \\
		15& learning rate & $0.01$, $0.001$, $0.0001$, $0.00001$\\
		\bottomrule			
	\end{tabular}
	
	\caption{Order in which different hyper-parameters are explored and the corresponding values considered for each hyper-parameter. Underlined values indicate the default value.}
	\label{tab0}
%	\begin{tablenotes}
%   \item[*] \footnotesize{Only used with inverse square root}  
%  \end{tablenotes}
% \end{threeparttable}
\end{table}

\subsection{Experimental setup}

Exploring all possible values for several hyper-parameters at once is prohibitively expensive from a computational perspective. 
Possible ways to circumvent this are random search \cite{BergstraB12} or grid search for one hyper-parameter at a time. 
For simplicity, we opt for the latter.
Table~\ref{tab0} shows the order in which the hyper-parameters are tuned. 
Once the optimal value of a hyper-parameter has been determined, it remains fixed for later steps; see Table~\ref{tab1}. 
Obviously, there are no guarantees that this will result in a global optimum.

To be comparable with \newcite{sennrich2019revisiting}, we take the TED data from the IWSLT 2014 German-English (De-En) shared translation task~\cite{cettolo2014report}. 
We apply punctuation normalization, tokenization, data cleaning, and truecasing using the Moses scripts~\cite{KoehnHBCFBCSMZDBCH07}. 
We also limit the sentence length to a maximum of 175 tokens during training. 
Our pre-processing pipeline results in 165,667 sentence pairs for training and 1,938 sentence pairs for development. 
In order to create smaller training sets, we randomly sample 5k, 10k, 20k, 40k, and 80k sentence pairs from the training data.
Similar to \newcite{sennrich2019revisiting}, we use the concatenation of the IWLST 2014 dev sets (tst2010--2012, dev2010, dev2012) as our test set, which consists of 6,750 sentence pairs.

%\textcolor{blue}{In order to tune the hyper-parameters, while commonly used approaches are  grid search and random search~\cite{BergstraB12}, we tune them step by step on top of each other, such that at each step the best evaluated point is picked, as shown in Table~\ref{tab0}. Obviously, each step is built on previous steps and if we change the order or pick another value for a hyperparameter at the current step, values of hyperparameters in the next steps can be slightly different. Note that at each step, all our parameter explorations are carried out on the development data, while results are only reported for the test sets.}

%\textcolor{blue}{As another approach in tunning related to our work, auto sizeing of model \cite{MurrayKNSC19} results in using less number of parameters in the encoder and decoder of the Transformer.}

For actual low-resource languages, we evaluate our optimized systems on the original test sets of Belarusian~(Be), Galician~(Gl), and Slovak~(Sk) TED talks~\cite{QiSFPN18} and also Slovenian~(Sl) from IWSLT2014~\cite{cettoloEtAl} with training sets ranging from 4.5k to 55k sentence pairs.
%As to actual low-resource languages, we evaluate our optimized systems on Slovenian$\rightarrow$English from IWSLT2014~\cite{cettoloEtAl}, which consists of 13k training examples, 2,181 parallel sentences for development and 2,555 for testing, and Slovak$\rightarrow$English from TED talks with 55k training examples, 2,271 sentence pairs for development, and 2,445 for testing.

We use Transformer-base and Transformer-big as our baselines, with the hyper-parameters and optimizer settings described in~\cite{vaswani2017attention}. 
We use the Fairseq library~\cite{ott2019fairseq} for our experiments and sacreBLEU~\cite{post-2018-call} as evaluation metric.

\subsection{Results and discussions}

\header{BPE effect} To evaluate the effect of different degrees of BPE segmentation on performance, we consider merge operations ranging from 1k to 30k, training BPE on the full training corpus instead of subsets and also removing infrequent subword units when applying the BPE model.
In contrast to earlier results for an RNN model, we observe that discarding infrequent subword units under extreme low-resource conditions is detrimental to the performance of Transformer. \newcite{sennrich2019revisiting} report that reducing BPE merge operations from 30k to 5k improves performance ($+4.9$ BLEU).  We find that the same reduction in merge operations affects the Transformer model far less~($+0.6$~BLEU). 
We observe no significant differences between training BPE on the full training corpus and training on subsets. Thus, we always train BPE on subsets with an optimized number of merge operations (see Table~\ref{tab2}).

\begin{table}[t]
	\centering
	\begin{tabular}{l lr | rr rrr}
		\toprule
		& & \multicolumn{6}{c}{BLEU}  \\ 
		\cmidrule{3-8}
		\multicolumn{1}{l}{ID} & \multicolumn{1}{l}{System} & \multicolumn{1}{c}{5k}   & \multicolumn{1}{c}{10k} & \multicolumn{1}{c}{20k}  & \multicolumn{1}{l}{40k} & \multicolumn{1}{c}{80k}  & 
		\multicolumn{1}{c}{165k} \\ 
		\midrule
		1 & Transformer-big &	$3.3 $ &	$3.4 $ &	$4.3$ &	$4.7$ &	$5.1$ & $5.5$ \\
		2 & Transformer-base&	$8.3$ &	$11.9$ &	$16.8$ &	$23.2$ &	$28.0$  & $32.1$\\
		\midrule
		3 & 2 + feed-forward dimension  (2048 $\rightarrow$ 512) &$8.8$ &	$12.0$	&$16.7$	&$22.3$& $27.7$ &$31.7$\\
		4&  3 +  attention heads (8$\rightarrow$  2)&$9.2$&	$12.7$&	$19.0$&	$23.6$& $28.7$ &$32.3$\\
		5& 4 + dropout (0.1$\rightarrow$   0.3 )&$10.6$&	$17.0$&	$21.9$&	$26.7$ & $\mathbf{31.0}$ &$\mathbf{33.4}$\\
		6  & 5 + layers (6 $\rightarrow$  5) &$10.9$	&	$16.9$&		$21.9$&		$26.0$& $30.2$	 &$33.0$\\
		7 & 6 + label smoothing (0.1$\rightarrow$  0.6)&$11.3$&	$16.5$	&$22.0$	&$26.9$ & $30.4$&$33.3$\\		
		8&  7 + decoder layerDrop (0 $\rightarrow$ 0.3) & {$12.9$}&	{$17.3$}&{$22.5$}&{$26.9$}& 30.3 & $33.1$\\
		9 & 8 + target word dropout (0 $\rightarrow$ 0.1)&$13.7$	&$18.1$	&$23.1$&	$27.0$ & $30.7$ &$33.0$\\
		10 & 9 + activation dropout (0 $\rightarrow$ 0.3)&$\mathbf{14.3}$	&$\mathbf{18.3}$&	$\mathbf{23.6}$&	$\mathbf{27.4}$&	$30.4$  &$32.6$\\	
		\bottomrule			
	\end{tabular}
	\caption{Results of Transformer optimized on the 5k dataset for different subsets and full corpus of IWSLT14 German $\rightarrow$ English. Averages over three runs from three different samples are reported.}
	\label{tab1}
\end{table}
\begin{table}[t]
	\centering
	%\small
	\begin{tabular}{l r| rrrrrr}
		\toprule
	     & default &5k & 10k & 20k & 40k & 80k & 165k \\ 
		\midrule
		BPE operations & $37$k & $5$k & $10$k &$10$k &$12$k &$15$k &$20$k \\
		feed-forward dimension & $2048$ & $512$ & $1024$ & $1024$ &$2048$ &$2048$ &$2048$ \\
		attention heads &$8$ & $2$& $2$ & $2$ & $2$ & $2$ & $4$ \\
		dropout &  $0.1$ & $0.3$ & $0.3$ & $0.3$& $0.3$&$0.3$ &$0.3$ \\
		layers & $6$ & $5$&$5$ &$5$ &$5$ &$5$ & $5$ \\
		label smoothing &  $0.1$ &$0.6$&$0.5$ &$0.5$ &$0.5$ &$0.4$ &$0.3$ \\	
		enc/dec layerDrop &$0.0/0.0$ &$0.0/0.3$ &$0.0/0.2$ &$0.0/0.2$ &$0.0/0.1$ &$0/0.1$ & $0/0.1$ \\
		src/tgt word dropout &$0.0/0.0$ & $0.0/0.1$ &$0.0/0.1$ &$0.1/0.1$ &$0.1/0.1$ &$0.2/0.2$ & $0.2/0.2$ \\
		activation dropout & $0.0$	& $0.3$ &$0.3$ &$0.3$ & $0.3$& $0$& $0$ \\
		batch size &$4096$ &$4096$ &$4096$ & $4096$&$4096$ & $8192$&$12288$ \\
		\bottomrule			
	\end{tabular}
	\caption{Default parameters for Transformer-base and optimal settings for different dataset sizes based on the De$\rightarrow$En development data.}
	\label{tab2}
	%\vspace{-0.4cm}
\end{table}

\header{Architecture effect} Table~\ref{tab1} shows the results of our system optimizations alongside the performance of our baselines.
We notice that Transformer-big performs poorly on all datasets, which is most likely due to the much larger number of parameters requiring substantially larger training data.
The system column in Table~\ref{tab1} shows our optimization steps on the 5k dataset, which are also applied to the larger datasets.  

We gain substantial improvements over Transformer-base for various subset sizes. 
For the smallest dataset, as expected, reducing Transformer depth and width, including number of attention heads, feed-forward dimension, and number of layers along with increasing the rate of different regularization techniques is highly effective ($+6$ BLEU). 
The largest improvements are obtained by increasing the dropout rate ($+1.4$ BLEU), adding layer dropout to the decoder ($+1.6$ BLEU), and adding word dropout to the target side ($+0.8$ BLEU). 
Most of these findings also hold for the 10k and 20k datasets, but differ for larger subsets.
By applying these settings to the 10k, 20k, 40k, 80k, and 165k datasets, BLEU scores increase by $+6.4$, $+6.8$, $+4.2$, $+2.4$, and $+0.5$ points, respectively. However, the effect of each adjustment is different for each dataset. For example, reducing the feed-forward layer dimension to $512$ is only effective for the two smallest subsets. 

We also conducted experiments with different values for the learning rate and warm-up steps using the inverse-square root learning rate scheduler, as implemented within Fairseq~\cite{ott2019fairseq}, which is slightly different from the proposed learning rate scheduler in the original Transformer paper. 
However, we did not observe any improvements over the default Transformer learning rate scheduler.
%that turns out to be not effective. Therefore, the standard learning rate scheduler for the Transformer is selected.}

%We further show that if we tune the parameters separately for each dataset, even more improvement is achieved. 
\def\@fnsymbol#1{\ensuremath{\ifcase#1\or *\or \dagger\or \ddagger\or
   \mathsection\or \mathparagraph\or \|\or **\or \dagger\dagger
   \or \ddagger\ddagger \else\@ctrerr\fi}}
\newcommand{\ssymbol}[1]{^{\@fnsymbol{#1}}}

\begin{table}[t]
\begin{minipage}{0.49\linewidth}
    \centering
    \begin{tabular}{lrrr}
     \toprule
     	& & \multicolumn{2}{c}{BLEU}  \\
     	\cmidrule{3-4}
     sentences & words (En) & T-base & T-opt  \\
     \midrule
     De$\rightarrow$En & & & \\
     $5$k & $100$k & $8.3$ & $14.3$ \\
     $10$k & $200$k & $11.9$ & $18.7$ \\
     $20$k & $410$k & $16.8$ & $24.1$ \\
     $40$k & $830$k & $23.2$ & $28.6$ \\
     $80$k & $1.6$M & $28.0$ & $31.9$ \\
     $165$k & $3.4$M & $32.1$ & $35.2$ \\
     \midrule
    %  Sl $\rightarrow$ En & & & \\
     Be$\rightarrow$En ($4.5$k)   & $90$k &$5.0$ & $8.1$\\
     Gl$\rightarrow$En ($10$k)  & $196$k  & $13.1$ & $22.3$ \\
     Sl$\rightarrow$En ($13$k) & $269$k & $9.1$ & $15.5$ \\
    %  \midrule
    %  Sk $\rightarrow$ En & & & \\
     Sk$\rightarrow$En ($55$k) & $1.2$M & $24.8$ & $29.9$ \\
     \bottomrule
    \end{tabular}
    \centering
    %\vspace{0.2cm}
	\caption{Results for Transformer-base/optimized. T-opt results for Be, Gl, Sl, and Sk use the optimized settings on De$\rightarrow$En development data for 5k, 10k, 10k, and 40k training examples, respectively.}
	%Results for Be$\rightarrow$En  data are reported using tokenized and lowercased test sets with \textit{multi-bleu.perl}
	\label{tab3}
\end{minipage}
\hspace{0.2cm}
\begin{minipage}{0.49\linewidth}
    \centering
    %\vspace{-0.1cm}
    \includegraphics[width=1\textwidth]{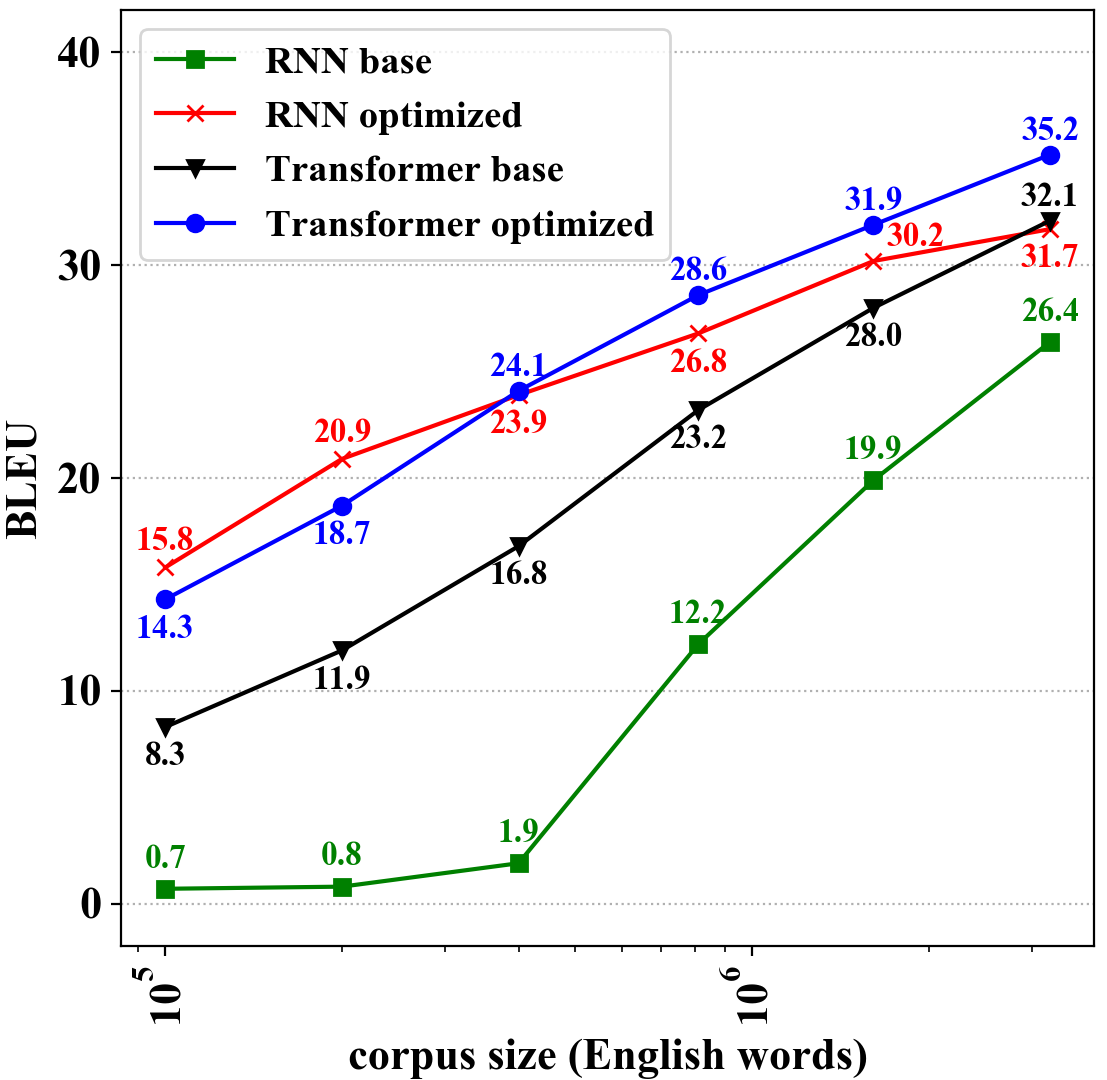}
    %\vspace{-0.6cm}
    \captionof{figure}{Comparison between RNN and Transformer with base and optimized settings.}, 
    %on different dataset sizes. }
    \label{fig:fig2}
\end{minipage}
%\vspace{-0.3cm}
\end{table}

\header{Optimized parameter settings} 
Table~\ref{tab2} shows the optimal settings for each dataset size, achieved by tuning the parameters on the development data.
We observe that a shallower Transformer combined with a smaller feed-forward layer dimension and BPE vocabulary size is more effective under lower-resource conditions. 
However, as mentioned above, Transformer is not as sensitive to the BPE vocabulary size as RNNs and reducing the embedding dimension size is not effective.
 
\newcite{vaswani2017attention} and \newcite{FosterVUMKFJSWB18} show that reducing the number of attention heads decreases the BLEU score under high-resource conditions. \newcite{Raganato} show that using one attention head does not cause  much degradation on moderate-size training data. Our results show that it is even beneficial to use only two attention heads ($+0.5$ BLEU) under low-resource conditions.
%, and four attention heads for the largest dataset size.
 
%Despite the fact that earlier findings~\cite{vaswani2017attention,FosterVUMKFJSWB18} under high-resource conditions demonstrate that reducing the number of attention heads decreases the BLEU score, \newcite{Raganato} show that using one attention head does not cause  much degradation on moderate-size training data. However, we find it beneficial to use two attention heads ($+0.5$ BLEU) under low-resource conditions, and four attention heads for the largest dataset size.

While \newcite{sennrich2019revisiting} use a high dropout rate of 0.5 for their optimized RNN model, our findings suggest a lower rate of 0.3 for Transformer. In line with their results, we find word dropout effective for most low-resource conditions. Our results show that a higher degree of label smoothing and higher decoder layer dropout rates are beneficial for smaller data sizes and less effective for larger sizes. 
 
%  Layer dropout has also been shown effective in high resource data conditions for very deep Transformers, with up to 18 layers~\cite{fan2019reducing}.

%Similar to \newcite{sennrich2019revisiting}, we also found word dropout effective for most low-resource conditions.

% Similar to RNNs for which word dropout is effective under both high-resource~\cite{SennrichHB16} and low-resource~\cite{sennrich2019revisiting} conditions, we also find smaller dropout rates are beneficial for most lower-resource conditions.
\newcite{sennrich2019revisiting} report substantial gains by using small batch sizes.
However, our results show that Transformer still requires larger batches, even under very low-resource conditions.
It is worth mentioning that applying attention dropout did not result in improvements  in our experiments.

\header{Optimized Transformer} 
The results of our optimized systems for the corresponding subsets are shown in the upper half of Table~\ref{tab3} with improvements of up to 3 BLEU points over the results obtained in Table~\ref{tab1}, indicating that under low-resource conditions, the optimal choice of Transformer parameters is highly sensitive with respect to the data size.
%The substantial differences between the results of Table~\ref{tab3} and Table~\ref{tab1} show that under low-resource conditions, the optimal choice of Transformer parameters does depend on the data size. 

The BLEU improvements in the bottom half of Table~\ref{tab3} show that determining the optimal settings on one language pair (De$\rightarrow$En) is also effective for actual low-resource language pairs, especially if the size of the training data is taken into account. Furthermore, the results in Table~\ref{tab:en2de} show that the optimal settings for De$\rightarrow$En also hold for the opposite translation direction of the same language pair.
They even carry over to translating from English for actual low-resource language pairs, see Table~\ref{tab:en2low}, which can be considered the more challenging scenario \cite{AharoniJF19}.
Note that the results of Tables~\ref{tab:en2de} and \ref{tab:en2low} and the bottom half of Table~\ref{tab3} are obtained by using the closest systems optimized on De$\rightarrow$En subsets with respect to their number of training sentences.

%our proposed optimizations are also effective for actual low-resource language pairs.

%\textcolor{blue}{Furthermore, while most of the time translating from English is more challenging than into English~\cite{AharoniJF19}, in Table~\ref{tab4}, we experimentally show that our obtained results also hold in the opposite translation directions. Note that the results of Table~\ref{tab4} and bottom half of Table~\ref{tab3} are obtained using the closest systems optimized on De$\rightarrow$En subsets with respect to their number of sentences.  }

% Belarusian, Galician, Slovenian and Slovak using systems optimized for 5k, 10k, 10k, and 40k training examples, respectively.} This confirms that the proposed optimizations are also effective for other low-resource language pairs.

To compare Transformer with an RNN architecture, we replicate the baseline and optimized RNN for low-resource NMT, as described in~\cite{sennrich2019revisiting}, on our datasets. Figure~\ref{fig:fig2} shows the BLEU scores for different data sizes. 
Surprisingly, even without any hyper-parameter optmization, Transformer performs much better than the RNN model under very limited data conditions. However, the optimized Transformer only outperforms the optimized RNN with more than 20k training examples.

%For this purpose, we take the four smallest datasets from En-De and our real low-resource datasets. As demonstrated in table\ref{tab4}, the optimized systems generalize on the inverse directions.  

\begin{table}[t]
   \begin{minipage}[t]{0.48\linewidth}
    \centering
    \begin{tabular}{lrrr}
     \toprule
     	& & \multicolumn{2}{c}{BLEU}  \\
     	\cmidrule{3-4}
     sentences & words (En) & T-base & T-opt  \\
     \midrule
     En$\rightarrow$De & & & \\
     $5$k & $100$k & $6.4$ & $11.3$ \\
     $10$k & $200$k & $9.3$ & $15.6$ \\
     $20$k & $410$k & $13.5$ & $20.8$ \\
     $40$k & $830$k & $20.2$ & $24.5$ \\
     $80$k & $1.6$M & $24.2$ & $27.2$ \\
     $165$k & $3.4$M & $27.4$ & $29.8$ \\
     \bottomrule
         \end{tabular}
\caption{\label{tab:en2de}Results for En$\rightarrow$De based on the optimal settings for De$\rightarrow$En for the corresponding corpus size (see Table~\ref{tab2}).}
% of translating from English using the exact same optimized models in Table~\ref{tab3}.}
	%Results for Be$\rightarrow$En  data are reported using tokenized and lowercased test sets with \textit{multi-bleu.perl}
	\label{tab4}
     \end{minipage}
  \hspace{0.2cm}
   \raisebox{0.725cm}[-0.725cm]{\begin{minipage}[t]{0.48\linewidth}
    \centering
    \begin{tabular}{lrrr}
     \toprule
     	& & \multicolumn{2}{c}{BLEU}  \\
     	\cmidrule{3-4}
     sentences & words (En) & T-base & T-opt  \\
     \midrule
    %  Sl $\rightarrow$ En & & & \\
     En$\rightarrow$Be ($4.5$k)   & $90$k &$3.6$ & $6.6$\\
     En$\rightarrow$Gl ($10$k)  & $196$k  & $10.6$ & $18.7$ \\
     En$\rightarrow$Sl ($13$k) & $269$k & $6.8$ & $12.2$ \\
    %  \midrule
    %  Sk $\rightarrow$ En & & & \\
     En$\rightarrow$Sk ($55$k) & $1.2$M & $19.5$ & $23.5$ \\
     \bottomrule
    \end{tabular}
    \centering
    %\vspace{0.2cm}
	\caption{\label{tab:en2low}Results for low-resource translation from English using the optimal settings from the De$\rightarrow$En system with the closest number of parallel sentence pairs.}
	%Results for Be$\rightarrow$En  data are reported using tokenized and lowercased test sets with \textit{multi-bleu.perl}
	\end{minipage}}
\end{table}

% \begin{table}[t]
%     \centering
%     \begin{tabular}{lrrr}
%      \toprule
%      	& & \multicolumn{2}{c}{BLEU}  \\
%      	\cmidrule{3-4}
%      sentences & words (En) & T-base & T-opt  \\
%      \midrule
%      En$\rightarrow$De & & & \\
%      $5$k & $100$k & $6.4$ & $11.3$ \\
%      $10$k & $200$k & $9.3$ & $15.6$ \\
%      $20$k & $410$k & $13.5$ & $20.8$ \\
%      $40$k & $830$k & $20.2$ & $24.5$ \\
%      $80$k & $1.6$M & $24.2$ & $27.2$ \\
%      $165$k & $3.4$M & $27.4$ & $29.8$ \\
%      \midrule
%     %  Sl $\rightarrow$ En & & & \\
%      En$\rightarrow$Be ($4.5$k)   & $90$k &$3.6$ & $6.6$\\
%      En$\rightarrow$Gl ($10$k)  & $196$k  & $10.6$ & $18.7$ \\
%      En$\rightarrow$Sl ($13$k) & $269$k & $6.8$ & $12.2$ \\
%     %  \midrule
%     %  Sk $\rightarrow$ En & & & \\
%      En$\rightarrow$Sk ($55$k) & $1.2$M & $19.5$ & $23.5$ \\
%      \bottomrule
%     \end{tabular}
%     \centering
%     %\vspace{0.2cm}
% 	\caption{\textcolor{blue}{Results of translating from English using the exact same optimized models in Table~\ref{tab3}.}}
% 	%Results for Be$\rightarrow$En  data are reported using tokenized and lowercased test sets with \textit{multi-bleu.perl}
% 	\label{tabX}
% \end{table}

\section{Conclusion}

In this paper, we study the effects of hyper-parameter settings for the Transformer architecture under various low-resource data conditions.
While our findings are largely in line with previous work \cite{sennrich2019revisiting} for RNN-based models, we show that very effective optimizations for RNN-based models such as reducing the number of BPE merge operations or using small batch sizes are less effective or even hurt performance. 
Our experiments show that a proper combination of Transformer configurations combined with regularization techniques results in substantial improvements over a Transformer system with default settings for all low-resource data sizes. However, under extremely low-resource conditions an optimized RNN model still outperforms Transformer.

% . While the results of an optimized Transformer model significantly improve for all low-resource data sizes, under extremely low-resource conditions an optimized RNN model still outperforms Transformer.

\newpage

\bibliography{main}

\begin{thebibliography}{}

\bibitem[\protect\citename{Aharoni \bgroup et al.\egroup }2019]{AharoniJF19}
Roee Aharoni, Melvin Johnson, and Orhan Firat.
\newblock 2019.
\newblock Massively multilingual neural machine translation.
\newblock In {\em Proceedings of the 2019 Conference of the North American
  Chapter of the Association for Computational Linguistics: Human Language
  Technologies, {NAACL-HLT} 2019}, pages 3874--3884.

\bibitem[\protect\citename{Bahdanau \bgroup et al.\egroup
  }2015]{bahdanau2014neural}
Dzmitry Bahdanau, Kyunghyun Cho, and Yoshua Bengio.
\newblock 2015.
\newblock Neural machine translation by jointly learning to align and
  translate.
\newblock In {\em 3rd International Conference on Learning Representations,
  {ICLR} 2015}.

\bibitem[\protect\citename{Bergstra and Bengio}2012]{BergstraB12}
James Bergstra and Yoshua Bengio.
\newblock 2012.
\newblock Random search for hyper-parameter optimization.
\newblock {\em The Journal of Machine Learning Research}, 13(1):281--305.

\bibitem[\protect\citename{Biljon \bgroup et al.\egroup }2020]{abs-2004-04418}
Elan~Van Biljon, Arnu Pretorius, and Julia Kreutzer.
\newblock 2020.
\newblock On optimal transformer depth for low-resource language translation.
\newblock {\em CoRR}, abs/2004.04418.

\bibitem[\protect\citename{Cettolo \bgroup et al.\egroup }2012]{cettoloEtAl}
Mauro Cettolo, Christian Girardi, and Marcello Federico.
\newblock 2012.
\newblock Wit3: Web inventory of transcribed and translated talks.
\newblock In {\em Conference of European Association for Machine Translation},
  pages 261--268.

\bibitem[\protect\citename{Cettolo \bgroup et al.\egroup
  }2014]{cettolo2014report}
Mauro Cettolo, Jan Niehues, Sebastian St{\"u}ker, Luisa Bentivogli, and
  Marcello Federico.
\newblock 2014.
\newblock Report on the 11th iwslt evaluation campaign, iwslt 2014.
\newblock In {\em Proceedings of the International Workshop on Spoken Language
  Translation, Hanoi, Vietnam}.

\bibitem[\protect\citename{Chen \bgroup et al.\egroup }2018]{FosterVUMKFJSWB18}
Mia~Xu Chen, Orhan Firat, Ankur Bapna, Melvin Johnson, Wolfgang Macherey,
  George~F. Foster, Llion Jones, Mike Schuster, Noam Shazeer, Niki Parmar,
  Ashish Vaswani, Jakob Uszkoreit, Lukasz Kaiser, Zhifeng Chen, Yonghui Wu, and
  Macduff Hughes.
\newblock 2018.
\newblock The best of both worlds: Combining recent advances in neural machine
  translation.
\newblock In {\em Proceedings of the 56th Annual Meeting of the Association for
  Computational Linguistics, {ACL} 2018}, pages 76--86.

\bibitem[\protect\citename{Cho \bgroup et al.\egroup }2014]{cho2014learning}
Kyunghyun Cho, Bart van Merrienboer, {\c{C}}aglar G{\"{u}}l{\c{c}}ehre, Dzmitry
  Bahdanau, Fethi Bougares, Holger Schwenk, and Yoshua Bengio.
\newblock 2014.
\newblock Learning phrase representations using {RNN} encoder-decoder for
  statistical machine translation.
\newblock In {\em Proceedings of the 2014 Conference on Empirical Methods in
  Natural Language Processing, {EMNLP} 2014}, pages 1724--1734.

\bibitem[\protect\citename{Fadaee \bgroup et al.\egroup }2017]{FadaeeBM17a}
Marzieh Fadaee, Arianna Bisazza, and Christof Monz.
\newblock 2017.
\newblock Data augmentation for low-resource neural machine translation.
\newblock In {\em Proceedings of the 55th Annual Meeting of the Association for
  Computational Linguistics, {ACL} 2017}, pages 567--573.

\bibitem[\protect\citename{Fan \bgroup et al.\egroup }2020]{fan2019reducing}
Angela Fan, Edouard Grave, and Armand Joulin.
\newblock 2020.
\newblock Reducing transformer depth on demand with structured dropout.
\newblock In {\em 8th International Conference on Learning Representations,
  {ICLR} 2020}.

\bibitem[\protect\citename{Gal and Ghahramani}2016]{gal2016theoretically}
Yarin Gal and Zoubin Ghahramani.
\newblock 2016.
\newblock A theoretically grounded application of dropout in recurrent neural
  networks.
\newblock In {\em Advances in Neural Information Processing Systems 29: Annual
  Conference on Neural Information Processing Systems}, pages 1019--1027.

\bibitem[\protect\citename{Gu \bgroup et al.\egroup }2018a]{GuHDL18}
Jiatao Gu, Hany Hassan, Jacob Devlin, and Victor O.~K. Li.
\newblock 2018a.
\newblock Universal neural machine translation for extremely low resource
  languages.
\newblock In {\em Proceedings of the 2018 Conference of the North American
  Chapter of the Association for Computational Linguistics: Human Language
  Technologies, {NAACL-HLT} 2018}, pages 344--354.

\bibitem[\protect\citename{Gu \bgroup et al.\egroup }2018b]{GuWCLC18}
Jiatao Gu, Yong Wang, Yun Chen, Victor O.~K. Li, and Kyunghyun Cho.
\newblock 2018b.
\newblock Meta-learning for low-resource neural machine translation.
\newblock In {\em Proceedings of the 2018 Conference on Empirical Methods in
  Natural Language Processing}, pages 3622--3631.

\bibitem[\protect\citename{G{\"{u}}l{\c{c}}ehre \bgroup et al.\egroup
  }2015]{gulcehre2015using}
{\c{C}}aglar G{\"{u}}l{\c{c}}ehre, Orhan Firat, Kelvin Xu, Kyunghyun Cho,
  Lo{\"{\i}}c Barrault, Huei{-}Chi Lin, Fethi Bougares, Holger Schwenk, and
  Yoshua Bengio.
\newblock 2015.
\newblock On using monolingual corpora in neural machine translation.
\newblock {\em CoRR}, abs/1503.03535.

\bibitem[\protect\citename{He \bgroup et al.\egroup }2016]{he2016dual}
Di~He, Yingce Xia, Tao Qin, Liwei Wang, Nenghai Yu, Tie{-}Yan Liu, and
  Wei{-}Ying Ma.
\newblock 2016.
\newblock Dual learning for machine translation.
\newblock In {\em Advances in Neural Information Processing Systems 29: Annual
  Conference on Neural Information Processing Systems}, pages 820--828.

\bibitem[\protect\citename{Koehn and Knowles}2017]{koehn2017six}
Philipp Koehn and Rebecca Knowles.
\newblock 2017.
\newblock Six challenges for neural machine translation.
\newblock In {\em Proceedings of the First Workshop on Neural Machine
  Translation, NMT@ACL 2017}, pages 28--39.

\bibitem[\protect\citename{Koehn \bgroup et al.\egroup
  }2007]{KoehnHBCFBCSMZDBCH07}
Philipp Koehn, Hieu Hoang, Alexandra Birch, Chris Callison{-}Burch, Marcello
  Federico, Nicola Bertoldi, Brooke Cowan, Wade Shen, Christine Moran, Richard
  Zens, Chris Dyer, Ondrej Bojar, Alexandra Constantin, and Evan Herbst.
\newblock 2007.
\newblock Moses: Open source toolkit for statistical machine translation.
\newblock In {\em {ACL} 2007, Proceedings of the 45th Annual Meeting of the
  Association for Computational Linguistics}.

\bibitem[\protect\citename{Konda \bgroup et al.\egroup
  }2015]{bouthillier2015dropout}
Kishore~Reddy Konda, Xavier Bouthillier, Roland Memisevic, and Pascal Vincent.
\newblock 2015.
\newblock Dropout as data augmentation.
\newblock {\em CoRR}, abs/1506.08700.

\bibitem[\protect\citename{M{\"{u}}ller \bgroup et al.\egroup
  }2019]{muller2019does}
Rafael M{\"{u}}ller, Simon Kornblith, and Geoffrey~E. Hinton.
\newblock 2019.
\newblock When does label smoothing help?
\newblock In {\em Advances in Neural Information Processing Systems 32: Annual
  Conference on Neural Information Processing Systems}, pages 4696--4705.

\bibitem[\protect\citename{Ott \bgroup et al.\egroup }2019]{ott2019fairseq}
Myle Ott, Sergey Edunov, Alexei Baevski, Angela Fan, Sam Gross, Nathan Ng,
  David Grangier, and Michael Auli.
\newblock 2019.
\newblock {Fairseq}: {A} fast, extensible toolkit for sequence modeling.
\newblock In {\em Proceedings of the 2019 Conference of the North American
  Chapter of the Association for Computational Linguistics: Human Language
  Technologies, {NAACL-HLT} 2019}, pages 48--53.

\bibitem[\protect\citename{Post}2018]{post-2018-call}
Matt Post.
\newblock 2018.
\newblock A call for clarity in reporting {BLEU} scores.
\newblock In {\em Proceedings of the Third Conference on Machine Translation:
  Research Papers, {WMT} 2018, Belgium, Brussels, October 31 - November 1,
  2018}, pages 186--191.

\bibitem[\protect\citename{Qi \bgroup et al.\egroup }2018]{QiSFPN18}
Ye~Qi, Devendra~Singh Sachan, Matthieu Felix, Sarguna Padmanabhan, and Graham
  Neubig.
\newblock 2018.
\newblock When and why are pre-trained word embeddings useful for neural
  machine translation?
\newblock In {\em Proceedings of the 2018 Conference of the North American
  Chapter of the Association for Computational Linguistics: Human Language
  Technologies, NAACL-HLT}, pages 529--535.

\bibitem[\protect\citename{Raffel \bgroup et al.\egroup
  }2019]{raffel2019exploring}
Colin Raffel, Noam Shazeer, Adam Roberts, Katherine Lee, Sharan Narang, Michael
  Matena, Yanqi Zhou, Wei Li, and Peter~J. Liu.
\newblock 2019.
\newblock Exploring the limits of transfer learning with a unified text-to-text
  transformer.
\newblock {\em CoRR}, abs/1910.10683.

\bibitem[\protect\citename{Raganato \bgroup et al.\egroup }2020]{Raganato}
Alessandro Raganato, Yves Scherrer, and J{\"{o}}rg Tiedemann.
\newblock 2020.
\newblock Fixed encoder self-attention patterns in transformer-based machine
  translation.
\newblock {\em CoRR}, abs/2002.10260.

\bibitem[\protect\citename{Sennrich and Zhang}2019]{sennrich2019revisiting}
Rico Sennrich and Biao Zhang.
\newblock 2019.
\newblock Revisiting low-resource neural machine translation: {A} case study.
\newblock In {\em Proceedings of the 57th Conference of the Association for
  Computational Linguistics, {ACL} 2019}, pages 211--221.

\bibitem[\protect\citename{Sennrich \bgroup et al.\egroup
  }2016]{sennrich2015improving}
Rico Sennrich, Barry Haddow, and Alexandra Birch.
\newblock 2016.
\newblock Improving neural machine translation models with monolingual data.
\newblock In {\em Proceedings of the 54th Annual Meeting of the Association for
  Computational Linguistics, {ACL} 2016}, pages 86--96.

\bibitem[\protect\citename{Sennrich \bgroup et al.\egroup
  }2017]{sennrich-etal-2017-nematus}
Rico Sennrich, Orhan Firat, Kyunghyun Cho, Alexandra Birch, Barry Haddow,
  Julian Hitschler, Marcin Junczys-Dowmunt, Samuel L{\"a}ubli, Antonio~Valerio
  Miceli~Barone, Jozef Mokry, and Maria N{\u{a}}dejde.
\newblock 2017.
\newblock {N}ematus: a toolkit for neural machine translation.
\newblock In {\em Proceedings of the Software Demonstrations of the 15th
  Conference of the {E}uropean Chapter of the Association for Computational
  Linguistics}, pages 65--68.

\bibitem[\protect\citename{Sutskever \bgroup et al.\egroup
  }2014]{sutskever2014sequence}
Ilya Sutskever, Oriol Vinyals, and Quoc~V. Le.
\newblock 2014.
\newblock Sequence to sequence learning with neural networks.
\newblock In {\em Advances in Neural Information Processing Systems 27: Annual
  Conference on Neural Information Processing Systems}, pages 3104--3112.

\bibitem[\protect\citename{Vaswani \bgroup et al.\egroup
  }2017]{vaswani2017attention}
Ashish Vaswani, Noam Shazeer, Niki Parmar, Jakob Uszkoreit, Llion Jones,
  Aidan~N. Gomez, Lukasz Kaiser, and Illia Polosukhin.
\newblock 2017.
\newblock Attention is all you need.
\newblock In {\em Advances in Neural Information Processing Systems 30: Annual
  Conference on Neural Information Processing Systems}, pages 5998--6008.

\bibitem[\protect\citename{Wang \bgroup et al.\egroup }2019]{wang2019learning}
Qiang Wang, Bei Li, Tong Xiao, Jingbo Zhu, Changliang Li, Derek~F. Wong, and
  Lidia~S. Chao.
\newblock 2019.
\newblock Learning deep transformer models for machine translation.
\newblock In {\em Proceedings of the 57th Conference of the Association for
  Computational Linguistics, {ACL} 2019}, pages 1810--1822.

\end{thebibliography}
\bibliographystyle{acl}

\end{document}